\documentclass{article}
\usepackage{spconf,amsmath,graphicx}
\usepackage{amsmath,amsfonts,amssymb}
\usepackage{lipsum} %

\usepackage[normalem]{ulem}
\usepackage{mathptmx} %
\usepackage{ifthen}
\usepackage{xspace}
\usepackage{graphicx}
\usepackage{tabu}
\usepackage{url}
\usepackage{tikz}
\usepackage{balance}

\newboolean{publicversion}
\setboolean{publicversion}{true}

\ifthenelse{\boolean{publicversion}}{
  \newcommand{\grumbler}[2]{}
  \newcommand{\grumblerPurple}[2]{}

}{
  \newcommand{\grumbler}[2]{\textcolor{blue}{\bf #1: #2}}
  \newcommand{\grumblerPurple}[2]{\textcolor{purple}{\bf #1: #2}}

  \usepackage{background}
  \backgroundsetup{position=current page.east, angle=-90, nodeanchor=east, vshift=-5mm, opacity=1, scale=2, contents=DRAFT for ICASSP 2020}
}

\newcommand{\squishlist}{
   \begin{list}{$\bullet$}
    { \setlength{\itemsep}{0pt}      \setlength{\parsep}{0pt}
      \setlength{\topsep}{0pt}       \setlength{\partopsep}{0pt}
      \setlength{\listparindent}{-2pt}
      \setlength{\itemindent}{-5pt}
      \setlength{\leftmargin}{1em} \setlength{\labelwidth}{0em}
      \setlength{\labelsep}{0.5em} } }

\newcommand{\squishend}{
    \end{list}  }

\usepackage{balance}

\newcommand{\fig}[1]{Fig.~\ref{#1}}

\newcommand{\langid}{LangID}

\title{Signal Combination for Language Identification}

\twoauthors
  {Shengye Wang\sthanks{The author did this work during his internship at Google. Email: \texttt{shengye@ucsd.edu}}}
	{University of California San Diego\\
	Computer Science and Engineering\\
	9500 Gilman Dr., La Jolla, CA 92093, USA}
  {Li Wan, Yang Yu, Ignacio Lopez Moreno\sthanks{Email: \texttt{\{liwan, yyuyy, elnota\}@google.com}}}
	{Google LLC., USA\\
	111 8th Ave., New York, NY 10011, USA
	}

\begin{document}
\maketitle
\begin{abstract}
Google's multilingual speech recognition system combines low-level acoustic signals with language-specific recognizer signals to better predict the language of an utterance.
This paper presents our experience with different signal combination methods to improve overall language identification accuracy.
We compare the performance of a lattice-based ensemble model and a deep neural network model to combine signals from recognizers with that of a baseline that only uses low-level acoustic signals.
Experimental results show that the deep neural network model outperforms the lattice-based ensemble model,
and it reduced the error rate from $5.5\%$ in the baseline to $4.3\%$, which is a $21.8\%$ relative reduction.

\end{abstract}

\section{Introduction}
\label{sec:intro}

\begin{figure}
\includegraphics[width=\columnwidth]{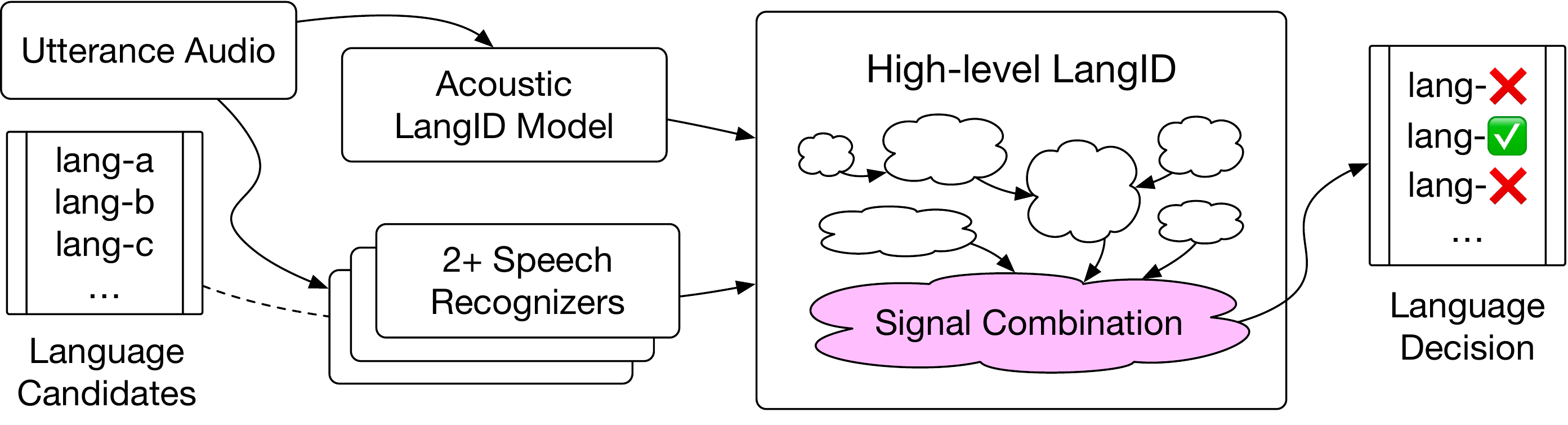}
\caption{\small{The ``\langid{}'' pipeline.
The system takes an audio clip of an utterance and a list of candidate languages, and it predicts the language of the utterance.}}
\label{fig:complete_pipeline}
\vspace{-2.5ex}
\end{figure}

Multilingual speech recognition is an important feature for modern speech recognition systems allowing users to speak in more than a single, preset language.
In Google multilingual speech recognition service~\cite{Gonzalez2015Multilingual} users are allowed to select two, or more, languages simultaneously as prior information~(\fig{fig:complete_pipeline}).
When the microphone is enabled, the system works by running several speech recognziers in parallel, along with an acoustic language identification (\langid{}) module~\cite{wan2019}.
After the system decides the language of the utterance, the recognition result of the corresponding language will be used and the language decision can be propagated to downstream systems (e.g. a text to speech module). 
In our previous work~\cite{Gonzalez2015Multilingual}, the final language identification decision is predominantly taken by the acoustic \langid{} module, which generates a probability score for each of the language based on the audio. Such approach however, ignores other potentially useful information returned by the individual speech recognizers such as the accumulated language or acoustic model score.

In this work, we explore a few alternative approaches to improve the language identification accuracy by using signals that can be easily computed by most speech recognition systems, including the confidence, acoustic and language model scores computed by the recognizers (Table \ref{tab:signals}).

Without signal combination, an acoustic \langid{} model provides a baseline with $5.5\%$ error rate.
Using a lattice-based ensemble model~\cite{Fard2016}, we were able to reduce the classification error rate to $4.5\%$, a relative $23.7\%$ reduction.
We continued exploring new methods and found a deep neural network model outperforms the lattice-based model: the error rate further reduced to $4.3\%$, which is a $21.8\%$ relative reduction from the original baseline.

This paper is structured as follows.
Section~\ref{sec:related-work} discusses related work in \langid{} and signal combination.
Section~\ref{sec:methods} presents our method with a lattice-based ensemble model and a deep neural network-based improvement.
We also explore methods in deep learning that work well in the signal combination problem.
Section~\ref{sec:results} shows the experimental results.
Finally, Section~\ref{sec:conclusion} concludes the paper.
\section{Related Work}
\label{sec:related-work}

Early work on speech-language identification traces back to 1996~\cite{Zissman1996Comparison}
where language identification is used on telephone speech.
Later, much work on language identification uses statistical learning tools such as Gaussian mixture models, support vector machines~\cite{Yang2007Tokenization, Campbell2008covariance, Ma2007Discriminative, Campbell2007Lattices,Zhu2008Discriminative}
and random forest based models~\cite{Wang2008forest}.
More recently, ~\langid{} system is a N-class acoustic-based classifier to generate scores for each language
~\cite{fer2015multilingual,snyder2018spoken,martinez2011language,wan2019}. So far the most successful types are neural network based methods~\cite{snyder2018spoken,wan2019}. They have been proven surpass the performance of the more traditional i-vector embeddings~\cite{martinez2011language}.

Combining signals form other system in addition to acoustics is a good way to further boost the performance of
~\langid{} accuracy, such as text-based features, language model features
~\cite{Sefara2016Text,Wang2019CNN}.
In this work, we have tried both lattice based methods and neural
network method to combine text-based semantic features and acoustic features to improve the accuracy of language identification. In our
experiments, the most optimal performance is achieved based
on our carefully designed neural network model.

\section{Methods}
\label{sec:methods}

\begin{table*}[ht]
\caption{
Input signals used in \langid{} signal combination.
}
\vspace{1em}
\label{tab:signals}
\centering
\begin{small}
\begin{tabular}{|l|p{140mm}|}
\hline
\bf Signal Name           & \bf Signal Descriptions                               \\ \hline
$langid\_score$       & The probability of the utterance among 79 different target languages~\cite{wan2019}. \newline The inputs are log-mel-filterbank of speech sequences and output is likelihood of each language.          \\ \hline
$am\_cost$            & The cost of the acoustic model from the recognizer.                  \\ \hline
$lm\_cost$            & The cost of the language model from the recognizer.                  \\ \hline
$confidence\_score$   & The confidence score of the recognition result.                 \\ \hline
$entropy\_score$      & The Levenshtein distance among every pair of hypotheses in the N best list (For $N=5$). The distance is computed at the character level using normalized hypotheses. \\ \hline
$text\_langid\_score$ & The language identification~\cite{zhang-etal-2018-fast-compact} score based on the recognized text.                \\ \hline
\end{tabular}
\end{small}
\end{table*}

We simplify the signal combination problem as binary classification by paring languages.
This section discusses the problem transformation, the data representation, and the classification models.

\subsection{Problem Formulation}

For each of the language candidates, the upstream systems generate six signals as shown in Table~\ref{tab:signals}~(one from the acoustic \langid{} model and five generated from recognizer results).
We pair different languages and concatenate the signals for language $\mathbf{a}$ and language $\mathbf{b}$
to turn the signal combination problem to a binary classification problem with $12$ inputs.
The goal is to make a classifier $f(\mathbf{a}, \mathbf{b})$ that outputs the probability of language $\mathbf{a}$ is preferred over language $\mathbf{b}$.

The classifier satifies a symmetry constraint: $f(\mathbf{b}, \mathbf{a}) = 1 - f(\mathbf{a}, \mathbf{b})$.
In other words, it should always generate opposite scores when exchanging languages $\mathbf{a}$ and $\mathbf{b}$.
When there are more than two language candidates, we pair all of them with each other and we assign a score $g(\mathbf{a}) = \sum_{\mathbf{a} \neq \mathbf{b}}{f(\mathbf{a}, \mathbf{b})}$ to language $\mathbf{a}$.
The language $\mathbf{a}$ with maximum $g(\mathbf{a})$ is selected as the outcome.
While there are other ways to rank all of the languages,
turning the problem into binary classification allows us to focus on the two-language scenario,
as there are more bi-lingual users than multi-lingual speakers.

\subsection{Dataset}
\label{sec:method:dataset}

We generated the training and testing datasets by randomly picking anonymized queries from Google Voice Search in 20 different languages.
Training and evaluation data was collected from monolingual traffic, from which we can infer the language labels in an unsupervised fashion.
We pair the signals from the correct language with signals from other languages, and we set the label to match the correct language.
Our dataset for the experiments contains $3M$ samples, which we divided into training and testing sets at $80\%:20\%$ ratio;
we split the dataset by the original utterance to prevent crosstalk between training and testing datasets.

Following the setting of our previous work~\cite{wan2018}, we use the average pairwise accuracy among all the language pairs to measure the system performance.
To reduce the computational cost of running evaluations on all the possible language pairs, in this paper we considered the performance metric to be the average pairwise accuracy among the top 15 language pairs as determined by the volume of traffic. 

\begin{figure}
\includegraphics[width=\columnwidth]{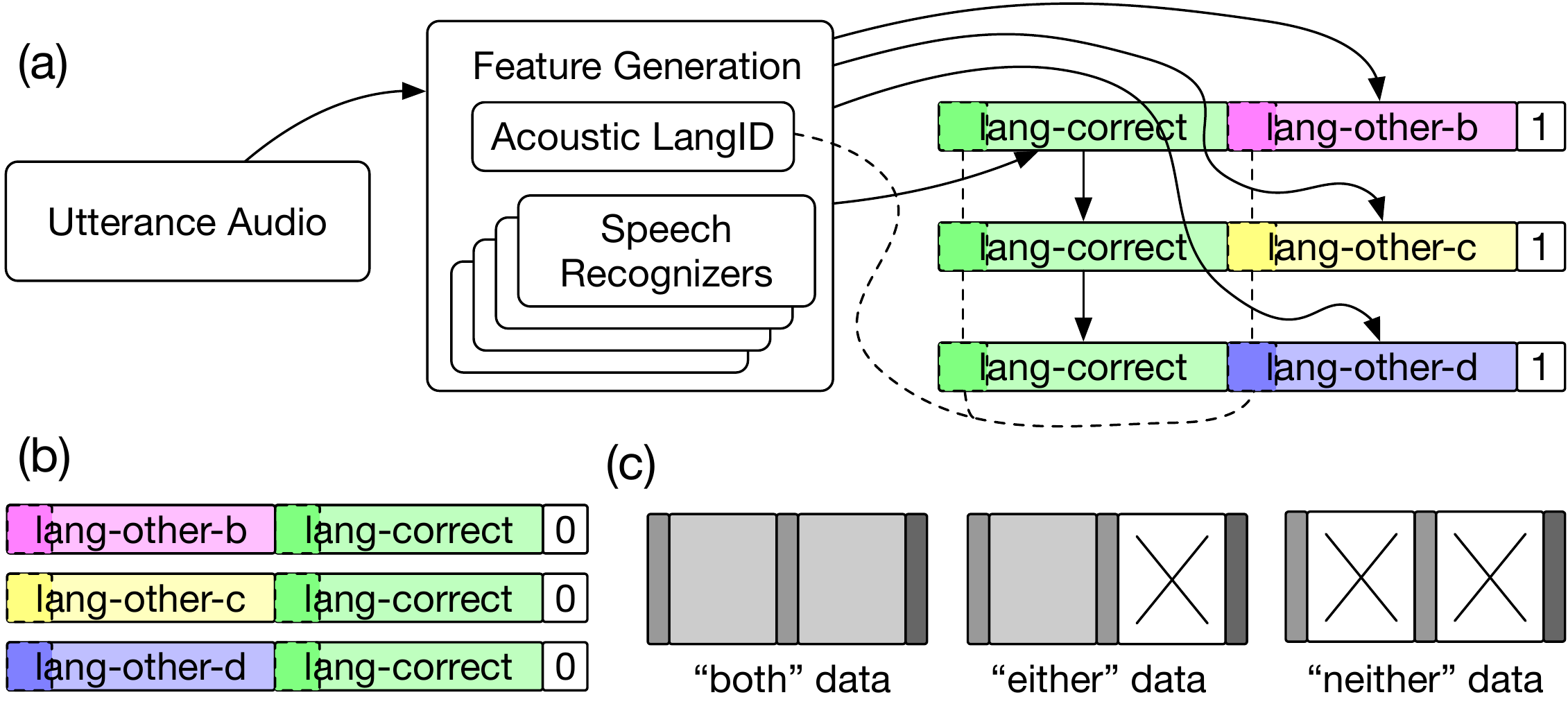}
\caption{\small{The data preparation process. (a)~The acoustic \langid{} model generates a $langid\_score$ for all concerned languages, and the recognizers and associated parts generate the other five features.\ (b)~The two sides are swapped to generate samples with negative labels.\ (c)~Not all recognizers finishes in time. Some of the samples only has 7 features ($langid\_score$-only on one side) or 2 features ($langid\_score$-only on both sides), while the others contain all $12$ features.}}
\label{fig:data_prep}
\vspace{-2.5ex}
\end{figure}

\fig{fig:data_prep} illustrates the dataset preparation process.
We run these samples through the acoustic \langid{} model~(The details of our acoustic \langid{} model is described in~\cite{wan2019}),
as well as recognizers of different languages to generate all the necessary signals.

\textbf{Missing Features}
Multiple speech recognizers run in parallel along with the acoustic \langid{} model.
Not all the language recognizers finish before the deadline to make a decision,
so the dataset contains missing features.
Some samples have all of the signals from both recognizers, and we refer these samples as ``both'' data.
Some samples have all of the signals from one of the recognizers,
but only acoustic \langid{} scores are available on the other side, and we refer these samples as ``either'' data.
For those samples with only acoustic \langid{} scores on both sides, we refer them as ``neither'' data.
Among the entire dataset,
$11.7\%$ are ``both'' data, $52.9\%$ are ``either'' data, and $35.4\%$ are ``neither'' data.

For the ``either'' dataset, we found a label bias towards the language with the recognizer outputs.
Among the ``either'' dataset, about $60\%$ of training samples has a label that matches the side with recognizer outputs, which is off from the $50\%$ expectation.
While whether the recognizer finishes in time is an important indication for classification,
we decided to model the bias in other parts of the pipeline.
We removed the bias from training data for signal combination by assigning higher weights to the samples has a label matches the no-recognizer-output side, and lower weights for the others.

As of the ``neither'' data, there is little we can do other than simply comparing the \langid{} scores.
The accuracy is the classification accuracy of the acoustic \langid{} model, and it is about $94.1\%$ as of now.
The next sections will discuss accurate predictions on the ``either'' and ``both'' data.
\subsection{Lattice Regression-based Ensemble Model}

Lattice regression with monotonic calibrations is a fast-to-evaluate and interpretable tool to combine signals to implement regression and classification~\cite{Gupta2016Monotonic}.
We implemented a lattice-based model and fine-tuned its parameters to combine signals for \langid{}.

The lattice regression model is an ensemble of $20$ sub-models.
The input features are calibrated using monotonic functions with linear inequality constraints.
Then, a sub-model only uses eight randomly-selected features from the $12$ calibrated inputs.
Our fine-tuning experiments conclude that slightly more emphasis on the acoustic model's scores ($langid\_score$) generates a higher accuracy --- we apply three lattices on $langid\_score$, and all the other features have two lattices.
Therefore, each sub-model is one of $2^8$, $3\times 2^7$, or $3^2 \times 2^6$ lattice functions, depending on whether $langid\_score$ from the two sides are used.

\textbf{Guaranteeing Symmetry}
One of the short-coming with the lattice-based model is lacking of symmetric.
The lattice-based model $f(\mathbf{a}, \mathbf{b})$ represents the probability that language $\mathbf{a}$ is preferred over language $\mathbf{b}$,
but when exchanging $\mathbf{a}$ and $\mathbf{b}$ it does not generate an opposite probability, i.e. 
for any $f$ based on lattice regression, $\exists\ \mathbf{a}, \mathbf{b}\ \text{s.t.}\ f(\mathbf{a}, \mathbf{b}) \neq 1 - f(\mathbf{b}, \mathbf{a})$.
Although the lattice model $f(\mathbf{a}, \mathbf{b})$ was trained with symmetric samples (for dataset $D$, $\forall (\mathbf{a}, \mathbf{b}, label) \in D, \exists (\mathbf{b}, \mathbf{a}, 1 - label) \in D$), the symmetry is not guarnateed during inference.
As a workaround, we compute $f^\prime(\mathbf{a}, \mathbf{b}) = \frac{f(\mathbf{a}, \mathbf{b}) + (1 - f(\mathbf{b}, \mathbf{a}))}{2}$ in prediction.

\subsection{Deep Neural Network Models}

While the lattice-based model is relatively usable, we run out of options to further improve the accuracy as there are only a few parameters to tune the model.
To further improve the accuracy of multi-language identification, we experiment with a deep-neural network.

\textbf{Symmetric Score Function}
As discussed in the previous section, one of the short-coming with the lattice-based model is lacking symmetry.
It has a workaround, but the workaround introduced inconsistency between training and inference. With more flexibility in a deep neural network (DNN), we designed an architecture that guarantees symmetry without differencing training and prediction.

\begin{figure}
\includegraphics[width=\columnwidth]{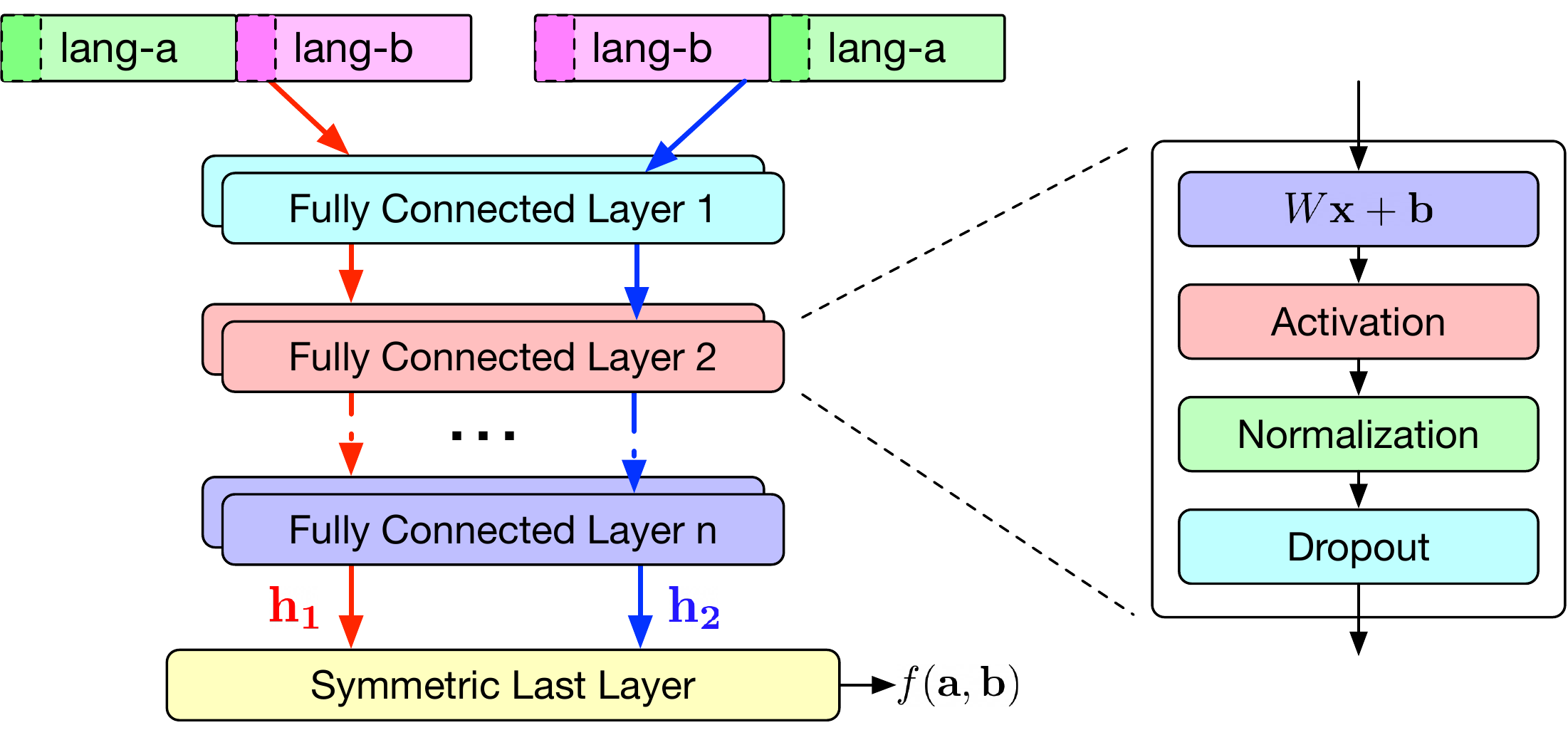}
\caption{\small{The deep neural network model architecture.
This architecture guarantees symmetry by having input $\mathbf{ab}$ and $\mathbf{ba}$ passing through the same network and combine them with a symmetric last layer.}}
\label{fig:dnn}
\vspace{-2.5ex}
\end{figure}

\fig{fig:dnn} shows the general architecture of our DNN model.
The input $\mathbf{ab}$ passes through a few fully-connected layers and generates $\mathbf{h_1}$;
then, the flipped input $\mathbf{ba}$ passes through the same layers and generates $\mathbf{h_2}$.
The loss function is defined as follows ($y=1$ when $a>b$, otherwise $y=0$):
$$
L(\mathbf{a},\mathbf{b},y)=y h(\mathbf{h_1}, \mathbf{h_2}) + (1-y) h(\mathbf{h_2}, \mathbf{h_1})
$$
Function $h$ defines how we combines the score $\mathbf{h_1}$ and $\mathbf{h_2}$. Experimentally, we find scaled cosine similarity function~\cite{wan2018} gives best performance:
$$
h(\mathbf{h_1}, \mathbf{h_2}) = sigmoid\left(w \frac{{\mathbf{h}_1}^T A^TA \mathbf{h_2}}{\|h1\|\|h_2\|}\right)
$$

\textbf{Network Architecture}
\fig{fig:dnn} also shows the elements in each of the fully connected layers: activation, dropout, and normalization.
We have tested different activation functions including $tanh$, $LeakyReLU$, and $ReLU$.
With only a few features in the network, dropouts improve model performance.
Normalization is another powerful tool to regularize the model.
We tested BatchNorm~\cite{BatchNormalization} and LayerNorm~\cite{ba2016layer} to stabilizes the network performance.
We also tested a different number of layers and number of elements of each layer including ``128-64-32-16-8'', ``64-64-64-64-64'', and residual connection~\cite{he2015deep} between them.

\textbf{Further Improve the Accuracy}
There are a few tools in deep neural networks that have the potential to improve the accuracy, just to name a few:
voting (model ensemble), parameter smoothing, and data augmentation.

An ensemble model is a good way to ``boost'' the accuracy of existing models. We trained $11$ models using the same architecture, and we test two methods to combine the voting results. Majority voting considers the same weight for all the models; on the other hand, averaging probability scores on each of the models generate an even higher accuracy.

Parameter smoothing prevents the evaluation metrics from frequent oscillation. We applied the exponential moving average on all trainable parameters in the model. But we found that the accuracy sightly decreased after applying parameter smoothing.
Our dataset contains missing features (Section~\ref{sec:method:dataset}). During our experiments with missing features (``either'' dataset),
we found training on ``either'' and ``both'' dataset increases the testing accuracy on the ``both'' dataset.
This leads to a hypothesis that missing features increase the generalizability of the model.
We found only testing on the ``both'' dataset, training with augmented data by masking either side of ``both'' training set yields higher accuracy than training without augmented data.

\section{Experimental Results}
\label{sec:results}

We compared multiple model configurations to find the best model.
The lattice-regression model has a similar implementation to~\cite{tflattice}.
We implemented the deep neural network models with TensorFlow.
This section presents the performance of these models.

\textbf{Best DNN Model Configuration}
The inputs are re-scaled to the $[-1, 1]$ range with simple transformation:
we applied $t(x) = 2x - 1$ on features in the range of $[0, 1]$;
for the features with a larger span, like the model loss, we took $log$ on the features and normalized them to the $[-1, 1]$ range.
We used a nabla-shape architecture:
the number of the hidden units on each of the fully connected layers are ``12(input)-128-64-32-16-8.''
All of the layers use $ReLU$ activation, $50\%$ dropout, and layer normalization~\cite{ba2016layer}.
In the loss function, $A$ is initialized with identity matrix and $w$ is initialized to $1$.
All $W$ in the fully-connected layers are initialized using He initialization~\cite{he2015delving}; 
$\mathbf{b}$ are initalized with $0$.
We trained $11$ models using the same configuration and averages their probability output before making a prediction.
The batch-size in training is $128$;
we used Adam optimizer~\cite{kingma2014adam} in training with an initial base learning rate of $0.01$.
We decayed the base learning rate to $0.005$, $0.001$, $0.0005$, $0.0001$ at 10M, 20M, 30M, and 40M steps respectively.
We stop at 50M steps when the performance no longer changes.

Table~\ref{tab:results} shows the performance of the DNN model and the lattice-regression model compared with the baseline setting using the acoustic \langid{} scores only.
For the ``both'' data, the language identification error rate is reduced by $52.4\%$;
there is $25.9\%$ error rate reduction for ``either'' data; and unchanged for ``none'' data since there is no additional information for these data. The identification error rate is reduced by $21.8\%$ on the whole testing dataset.

\begin{table}[ht]
\vspace{-1em}
\caption{
\small{Error rate for different signal combination models. The baseline is \langid{} network only; Lattice and DNN
are different models for signal combination. The evaluation splits the data based on different feature missing conditions.
``both'' data contains recognizer features from both languages; ``either'' data contains recognizer features from only one language; and ``none'' data contains only \langid{} features.}
}
\vspace{0.5em}
\label{tab:results}
\centering
\begin{tabu} to \linewidth {|X[1,c,m]|X[1,c,m]|X[1,c,m]|X[1,c,m]|X[1,c,m]|X[1,c,m]|}
\hline
\bf Model  & \bf ``both'' & \bf ``either'' & \bf ``none'' & \bf ``all'' \\ \hline
Baseline & 4.2\% & 5.4\% & 5.9\% & 5.5\%       \\ \hline
Lattice  & 2.4\% & 4.0\% & 5.9\% & 4.5\%       \\ \hline
DNN      & 2.0\% & 4.0\% & 5.9\% & 4.3\%       \\ \hline
\end{tabu}
\vspace{-1em}
\end{table}

The models are further tested on the multilingual speech recognition system~\cite{Gonzalez2015Multilingual}. The system uses the model proposed in this paper to determine the spoken language and turns the utterance into text with the speech recognizer in the corresponding language when the languages are know to the system. The average word error rate is $10.9\%$. With the DNN model, we reduces the identification error rate to $3.3\%$ and achieves $12.8\%$ word error rate~(WER), which is $2.6\%$ better than the Lattice model.

\begin{table}[ht]
\vspace{-1em}
\caption{\small{The evaluation on our multilingual speech recognition system. The system will detect the language of the utterance and turn it into text. The language identification error and the word error rate averaged among supported languages are evaluated. The \langid{}-only baseline is compared with other algorithms. One special case is the oracle system that knows the spoken language in advance.}}
\vspace{0.5em}
\label{tab:wer}
\centering
\begin{tabular}{|c|c|c|}
\hline
\bf Model & \bf ID error & \bf WER \\ \hline
Baseline & 4.0\% & 13.6\% \\ \hline
Lattice & 3.6\% & $13.2\%$ \\ \hline
DNN & 3.3\% & $12.8\%$ \\ \hline
Oracle & - & 10.9\% \\ \hline
\end{tabular}
\vspace{-2em}
\end{table}

\section{Conclusion}
\label{sec:conclusion}

In this paper, we show deep neural network is a powerful tool for signal combination in the multi-language identification problem.
Experimental results show $21.8\%$ reduction in error rate.
The deep neural network model is easier to train and has more flexibility to fine-tune,
and it outperforms our fine-tuned lattice-based model.

We believe there are more potentials in using deep neural networks for the signal combination than using lattice regression.
Our future work is to further improve the accuracy of signal combination and support better multi-language speech recognition.

\newpage
\small{
\balance
\bibliographystyle{IEEEbib}
\bibliography{reference}
}
\end{document}